\documentclass{article}
\usepackage[numbers]{natbib}
\usepackage{censor}
\StopCensoring
\usepackage{spconf,amsmath,graphicx,booktabs,amssymb,subfigure,color}

\usepackage{tabularx}
\usepackage{multirow}
\usepackage{enumitem}

\title{BEV-VLM: Trajectory Planning via Unified BEV Abstraction}
\name{\censor{Guancheng Chen$^{1\dagger}$, Sheng Yang$^{1\dagger}$, Tong Zhan$^1$ and Jian Wang$^{1}$}}
\address{\censor{$^1$School of Data Science, Fudan University, Shanghai, China\\
\texttt{\{gcchen25, sheng\_yang21\}@m.fudan.edu.cn} \\
  \texttt{\{tongzhan, jian\_wang\}@fudan.edu.cn}}}

\begin{document}
%
\maketitle
\begin{abstract}
This paper introduces BEV-VLM, a novel approach for trajectory planning in autonomous driving that leverages Vision-Language Model (VLM) with Bird’s-Eye View (BEV) feature map as visual input. Unlike conventional trajectory planning approaches that rely solely on raw visual data (e.g., camera images), our method utilizes highly compressed and informative BEV representation generated by fusing camera and LiDAR data and aligning them with HD Map. Since the unified BEV-HD Map format offers a geometrically consistent and rich scene description, it enables VLM to perform accurate trajectory planning. 
Experimental results on nuScenes demonstrate that, compared with \textit{state-of-the-art} vision-only methods, our approach yields $53.1\%$ planning accuracy improvement and achieves \textit{complete collision avoidance}. Our work highlights that VLM can effectively interpret processed visual representation like BEV feature, expanding their applicability beyond raw images in trajectory planning.
\end{abstract}

\section{Introduction}
\label{sec:intro}

Recently, the pursuit of advanced autonomous driving has attracted extensive attention with Vision-Language Model (VLM) emerging as a promising pathway, owing to their inherent cognitive capabilities from pre-training that facilitate effective application in real-world scenarios. While existing research has demonstrated the feasibility and reliability of using VLM for path planning by feeding camera images, these approaches have two critical limitations: i) they depend only on camera data and lack integration with other modalities (e.g., LiDAR point clouds), and ii) they generally neglect exploring other forms of scene representation, where Bird’s-Eye View (BEV) feature serves as a superior scene descriptor.

In this paper, to address the above limitations, we avoid the direct use of raw visual signals (e.g., camera images) as VLM inputs, but instead leverage pretrained sensor-to-BEV feature models (e.g., BEVFusion~\citep{liu2022bevfusion}, LSS~\citep{philion2020lift}) and standard feature map generation techniques to preprocess the multi-modal sensor data. In doing so, we obtain highly compressed yet information-dense BEV feature map with three critical advantages: i) they condense key driving environment information into a single representation, reducing computational overhead, ii) provide a unified space for multi-sensor fusion that accommodates sensors with different intrinsic and extrinsic parameters, and iii) exhibit geometric consistency with readily available High-Definition (HD) Map. After aligning BEV feature with HD Map based on local position, we render the latter onto the former to form a unified BEV-HD Map, which, together with the road topological structure information, enables VLM to perform more accurate trajectory planning. We formulate our contributions as follows:

    \begin{itemize}
        \item We propose a novel approach for applying VLM to autonomous driving by adopting BEV feature map as structured visual input. This input provides a geometrically consistent representation of driving scenes, offering a more task‑informed visual signal.
        \item By converting 3D spatial reasoning into explicit 2D topological understanding, the highly compressed BEV-HD Map representation well integrates the multi-sensor data and HD Map, allowing the model to generate trajectories based on a globally consistent spatial context and achieve superior planning accuracy over perspective-view-based methods.
    \end{itemize}

\section{Related Works}
\label{sec:related_works}

\begin{figure*}[t]
    \centering
    \includegraphics[width=1.0\linewidth]{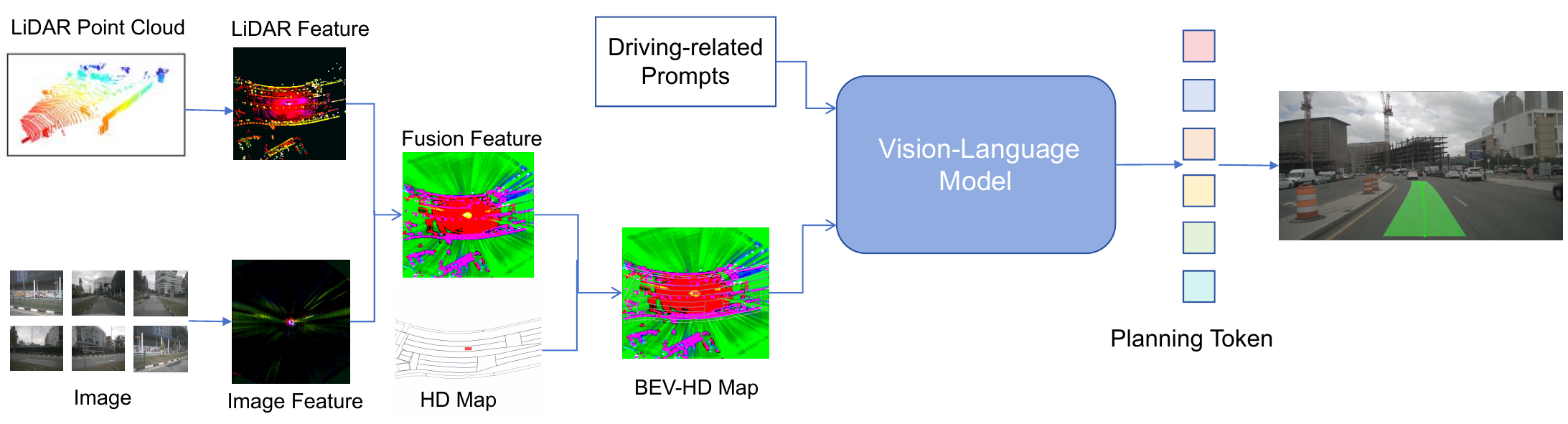}
    \caption{\textbf{Overall Architecture of BEV-VLM} BEV-VLM encompasses two core modules. It acquires BEV feature via pre-trained model, then integrates the BEV feature with HD Map to produce detailed scene description. The framework feeds the generated scene description into VLM, leveraging the capabilities of model to execute trajectory planning.}
    \label{fig:main}
\end{figure*}

\noindent \textbf{BEV Perception.} BEV feature has fully demonstrated their capability in representing spatiotemporal scenes. Works such as BEVFusion and BEVFormer~\citep{li2022bevformer} derived BEV feature by extracting information from multi-sensor data, achieving strong performance in perception tasks. BEVerse~\citep{zhang2022beverse} further extended the application of BEV feature to trajectory planning, validating their applicability in this domain.  

\noindent \textbf{BEV for Autonomous Driving.} In end-to-end autonomous driving methods, the BEV representation remains widely adopted in numerous works. ST-P3~\citep{hu2022stp3} addressed tasks like depth estimation and BEV segmentation by leveraging enhanced spatiotemporal BEV feature. UniAD~\citep{hu2023_uniad} employed a unified BEV representation to support multiple downstream tasks, including object detection and trajectory prediction. VAD~\citep{jiang2023vad, chen2024vadv2}, by contrast, converted BEV feature into structured vectorized representation via decoder, allowing BEV representation to directly contribute to planning, with vectorization constraints further enhancing safety.

\noindent \textbf{VLM for Autonomous Driving.} Boasting exceptional reasoning skills and abundant commonsense, large models are playing an increasingly vital role in autonomous driving. Drive-with-LLMs~\citep{chen2024drivingwithllms} generated future trajectory plans by feeding perception information in the latent space to large models. As a structured autonomous driving system, Senna~\citep{jiang2024senna} decoupled VLM-generated high-level planning decisions from end-to-end low-level trajectory predictions. VLM-AD~\citep{xu2024vlm} leveraged VLM as teacher to automatically generate annotations including unstructured reasoning text and structured action labels, integrates these annotations as supplementary supervisory signals. OmniDrive~\citep{wang2024omnidrive} presented a VLM for autonomous driving based on counterfactual reasoning, with a scalable pipeline for high-quality 3D driving Q\&A generation. AutoVLA~\citep{zhou2025autovla} integrated physical action tokens into a pretrained VLM backbone, enabling dual thinking modes through supervised fine-tuning. Max-V1~\citep{yang2025less} formulated autonomous driving trajectory planning as an end-to-end next waypoint prediction task based on a pure VLM. While the above  examples have demonstrated the reliability of VLM for trajectory prediction with the aforementioned information, direct research on trajectory prediction targeting BEV feature remains lacking.

\section{Methodology}

This work leverages VLM for trajectory planning based on BEV feature representation in autonomous driving scenarios, verifying that VLM can exploit the rich perceptual information of BEV feature for this task; the complete process is shown in Figure~\ref {fig:main}. We obtain BEV feature via a pre-trained BEV perception model, fuse the visualized BEV feature map with the HD Map to build a visual description of the ego-vehicle’s surroundings, and then feed these fused map directly into the VLM for trajectory planning.

\begin{figure*}[t]
    \centering
    \subfigure[Turn Left]{
        \includegraphics[width=0.235\linewidth]{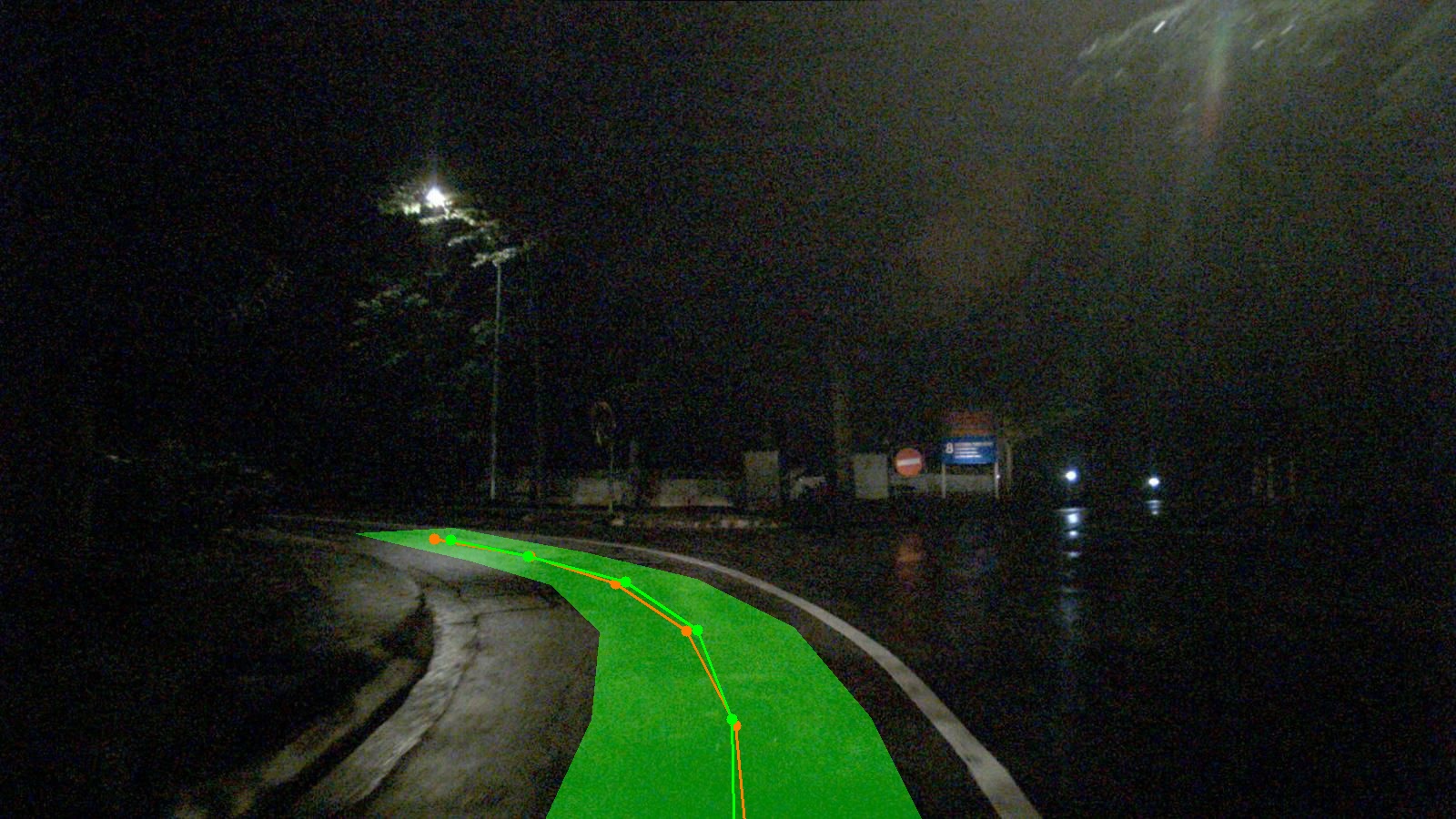}
        \label{fig:sub1_turnleft}
    }
    \subfigure[Go Straight]{
        \includegraphics[width=0.235\linewidth]{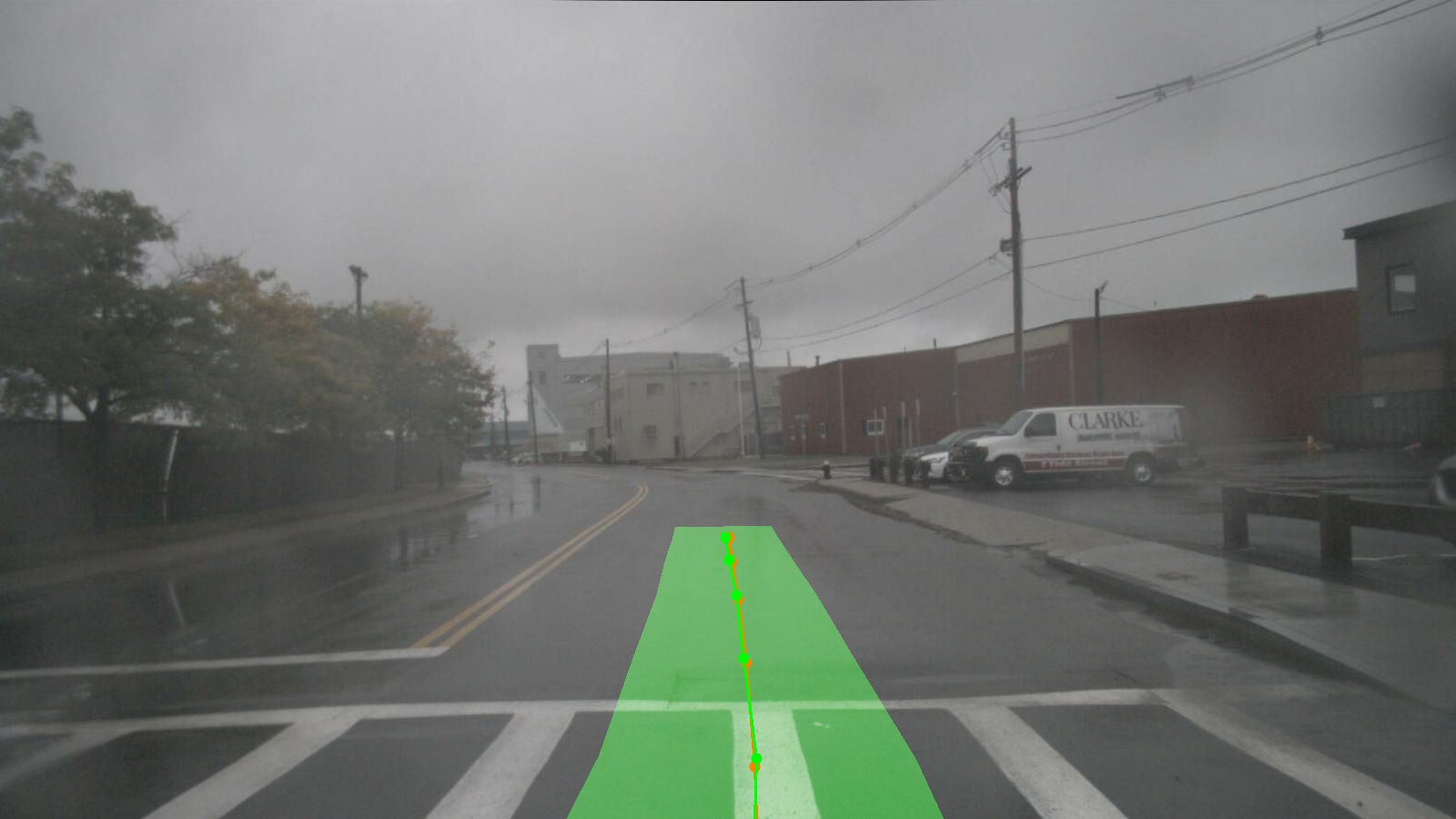}
        \label{fig:sub2_straight}
    }
    \subfigure[Prevent Collision]{
        \includegraphics[width=0.235\linewidth]{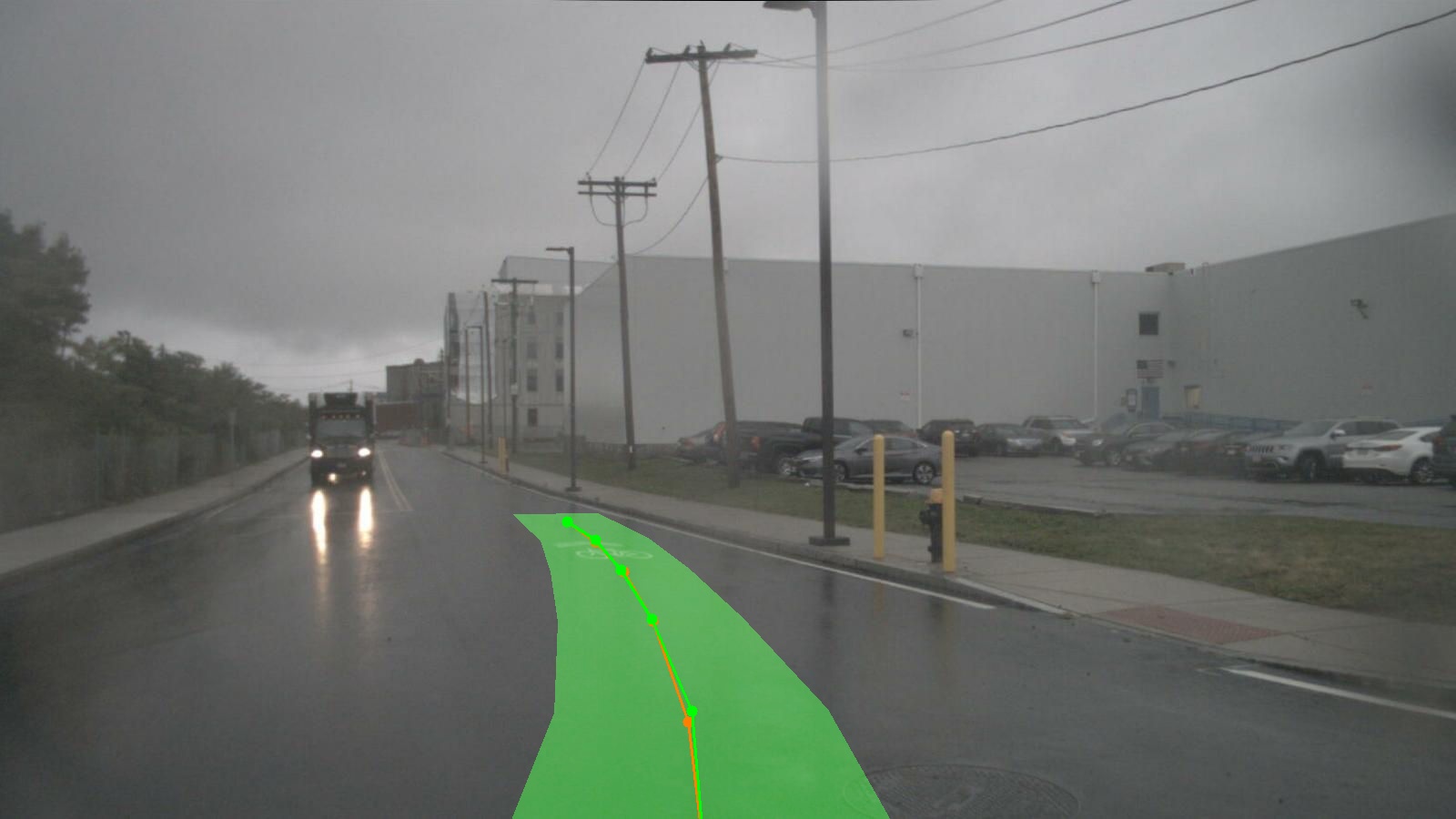}
        \label{fig:sub1_avoid}
    }
    \subfigure[Follow Car in Front]{
        \includegraphics[width=0.235\linewidth]{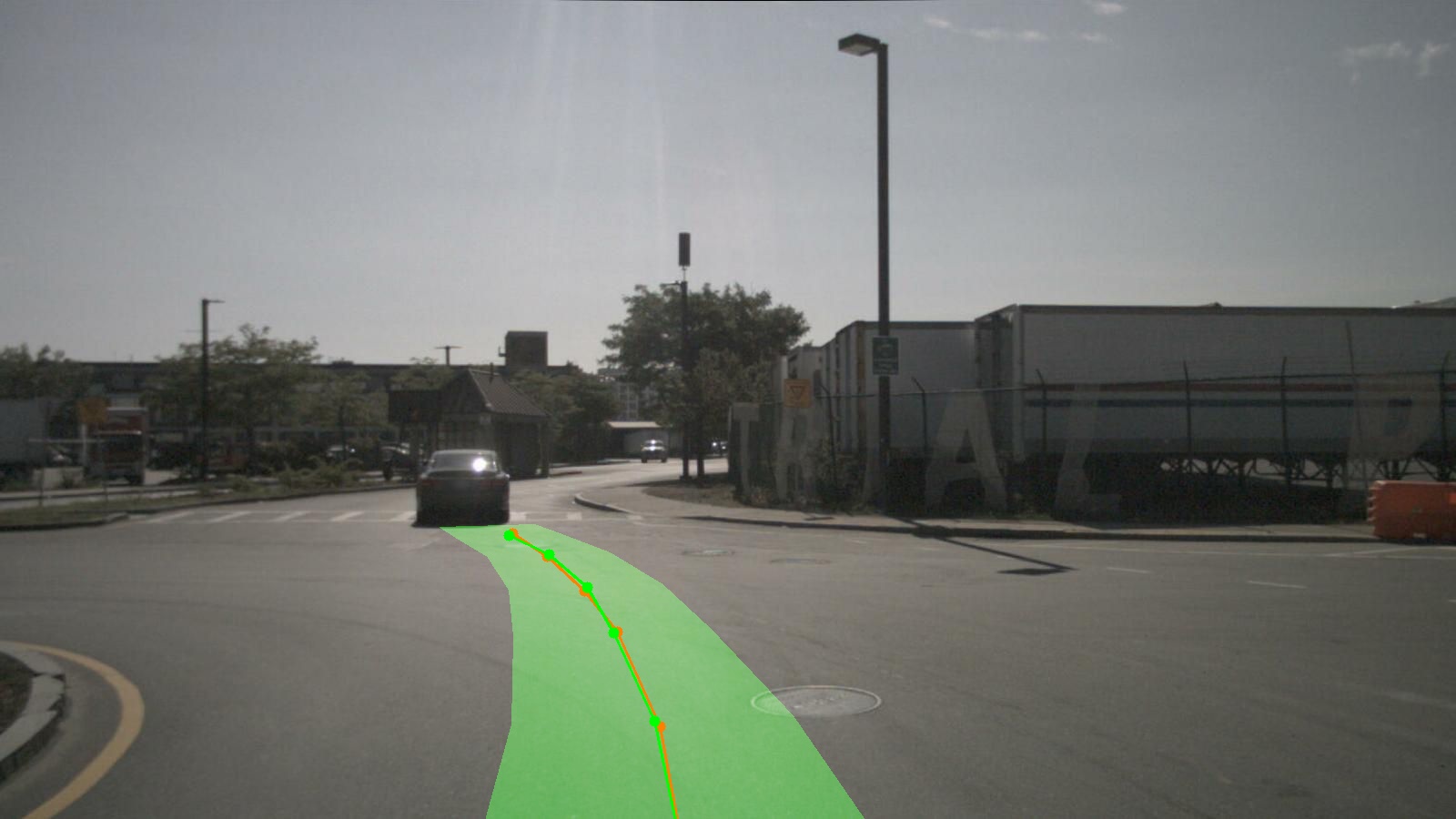}
        \label{fig:sub4_straight}
    }
    \subfigure[Turn Right]{
        \includegraphics[width=0.235\linewidth]{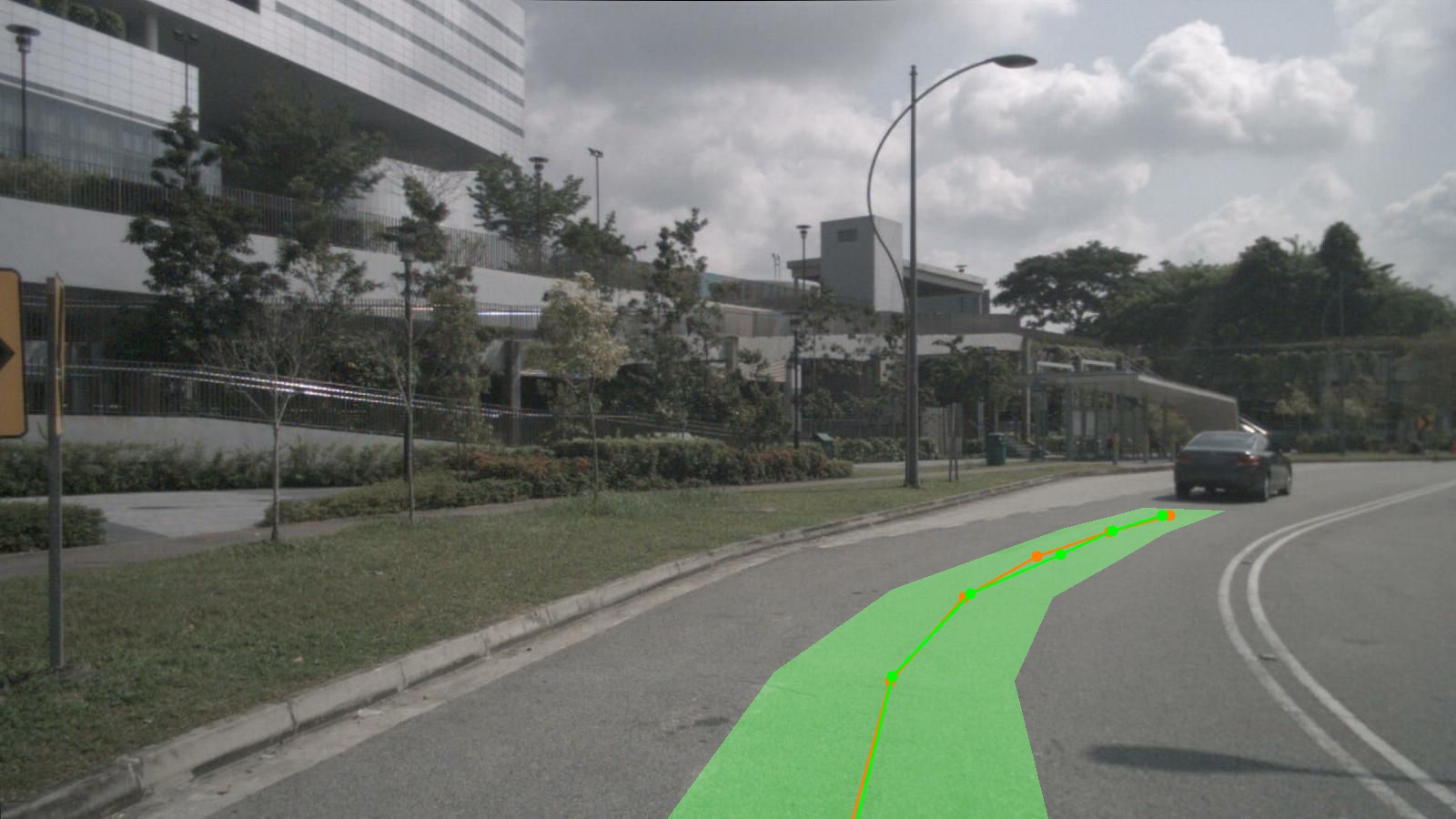}
        \label{fig:sub3_turnright}
    }
    \subfigure[Change Lanes]{
        \includegraphics[width=0.235\linewidth]{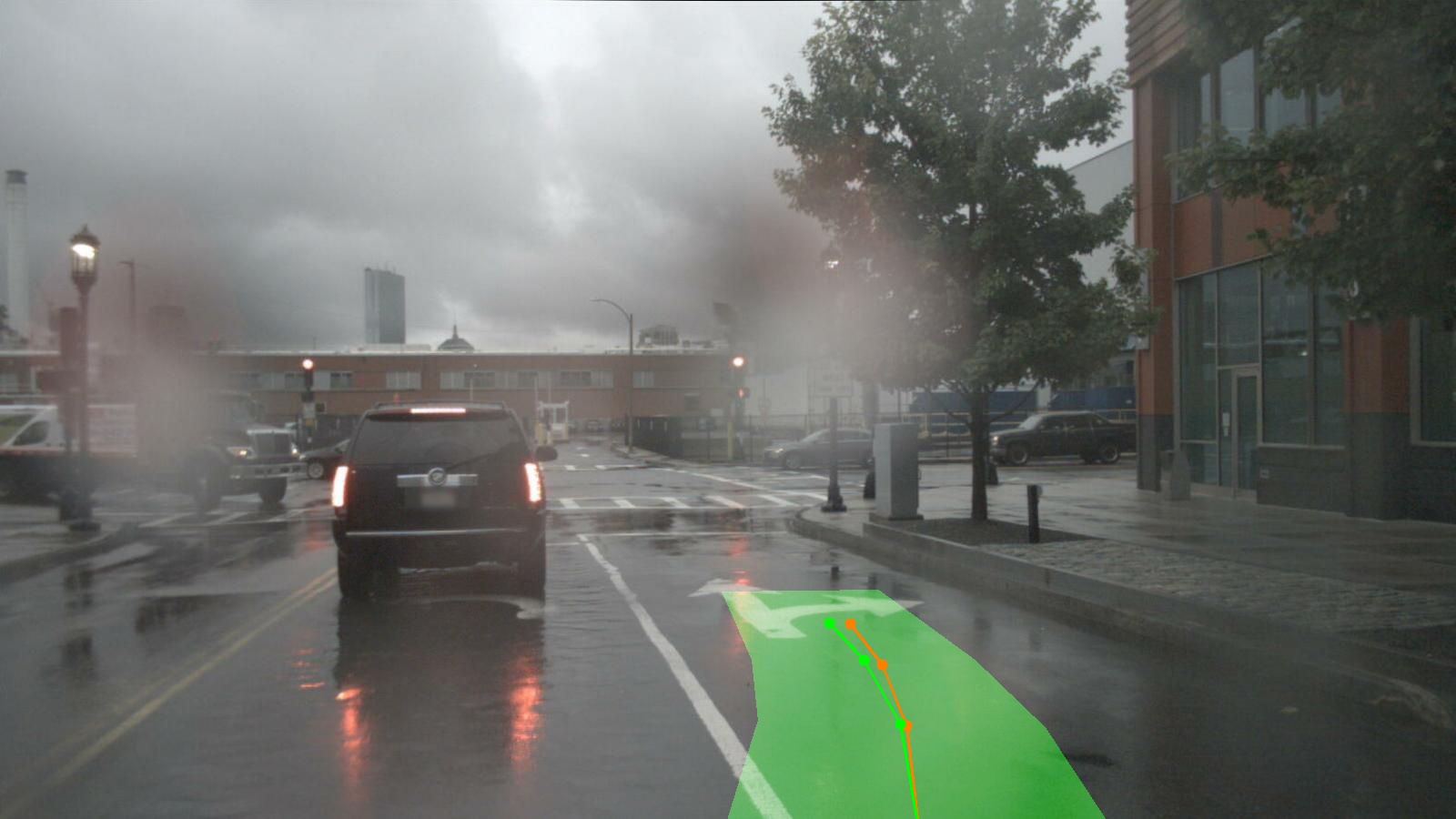}
        \label{fig:sub3_turnright}
    }
    \subfigure[Yield to Pedestrians]{
        \includegraphics[width=0.235\linewidth]{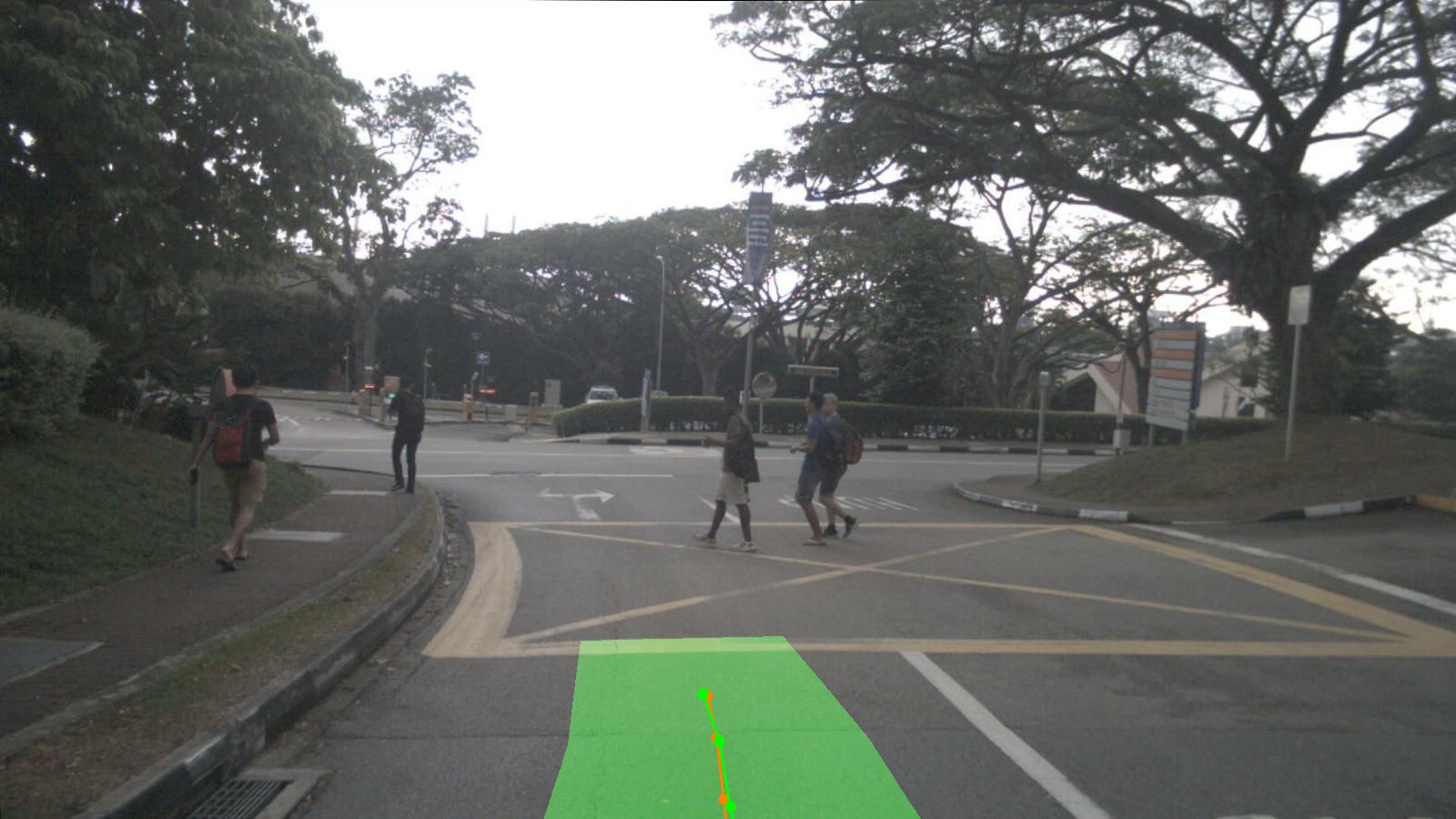}
        \label{fig:sub3_turnright}
    }
    \subfigure[Stop at Red Lights]{
        \includegraphics[width=0.235\linewidth]{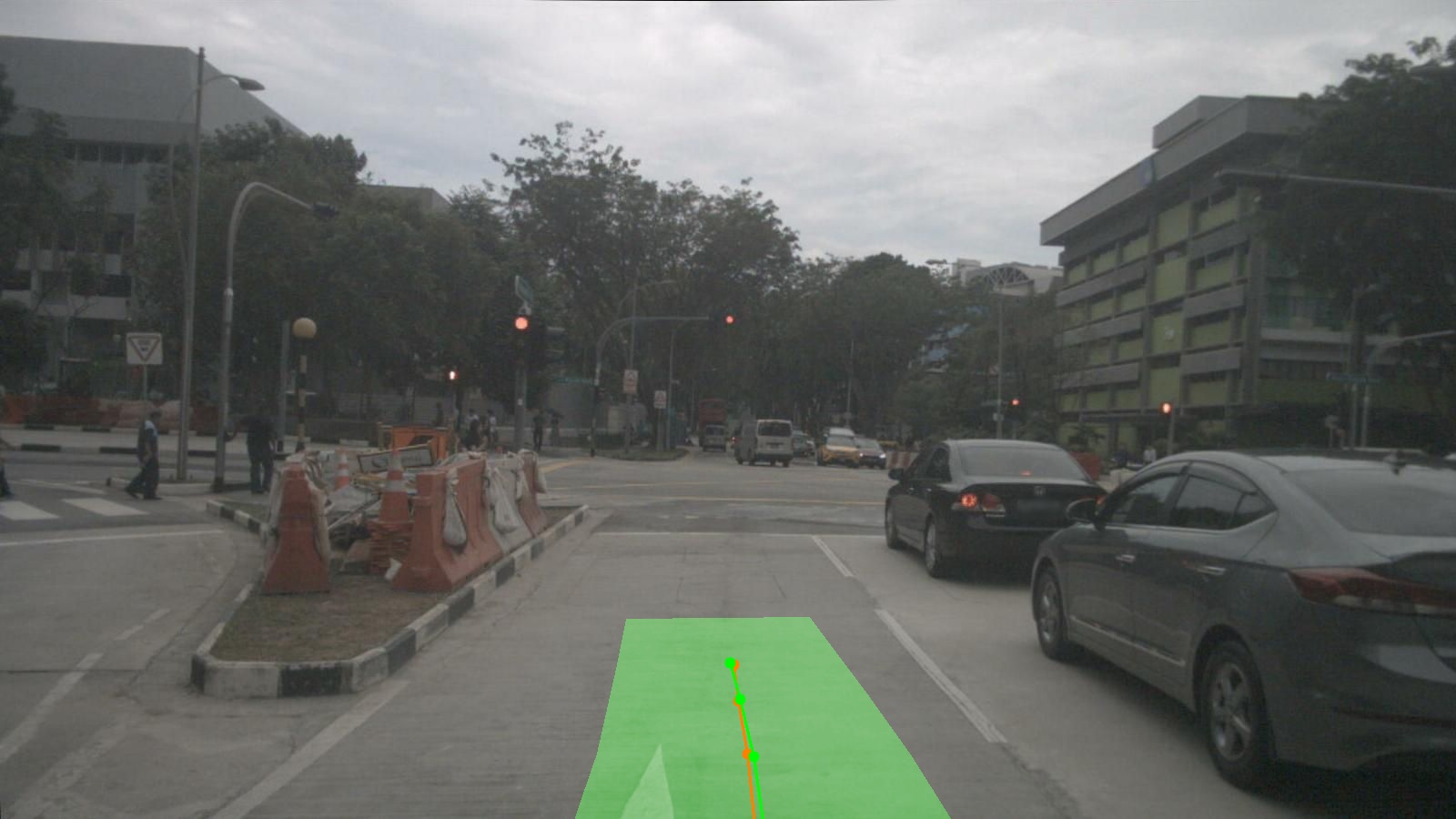}
        \label{fig:sub3_turnright}
    }
    \caption{\textbf{Cross-Column Visualization of Trajectories} 
    Several different scenarios are displayed, including various driving actions. Green denotes the predicted trajectories, while orange denotes the ground truth trajectories.}
    \label{fig:three_trajectories_crosscol}
\end{figure*}

\subsection{Scenario Description}

Existing VLM-based trajectory planning methods predominantly rely on raw visual inputs, such as multi-view images or video streams. However, these raw data formats introduce some challenges. First, they contain excessive task-irrelevant information, leading to a low signal-to-noise ratio for driving-specific reasoning. Second, processing and aligning high-resolution multiple perspectives input  imposes a prohibitive computational overhead on VLM. Specifically, standard VLM image encoders are typically pretrained on general computer vision benchmarks and lack the sensitivity required to perceive subtle changes in the direction of the ego-vehicle. In perspective-view images, a slight deviation in vehicle orientation may yield negligible changes at the pixel-level, but dictate a fundamentally different future trajectory. To address this, we propose to use a fused representation comprising the visualized BEV feature map and its corresponding HD Map as the primary scenario description. By projecting feature into the BEV space, the head of ego-vehicle is inherently aligned with the spatial coordinate system. Integrating HD Map further provides the VLM with explicit geometric constraints (e.g., lane-level topology). This approach harnesses the VLM’s spatial reasoning while preserving critical orientation cues and road geometry precisely.

The visualization of BEV feature for VLM input is designed as a 3-stage pipeline consisting of feature compression, ego-centric spatial normalization with self-vehicle alignment, and HD Map integration. First, high-dimensional BEV tensors $\mathcal{F} \in \mathbb{R}^{H \times W \times C}$ are dimensionally reduced via Principal Component Analysis (PCA), projecting the most discriminative driving feature into RGB representation compatible with standard VLM image encoders. Second, the BEV representation is spatially normalized to an ego-centric coordinate frame, with the self-vehicle anchored at the image center and its heading aligned with the rightward axis. This alignment encodes the rotational offset between the self-vehicle and road geometry as an explicit 2D planar transformation, thus yielding a spatially consistent representation that can be effectively captured by the VLM. Finally, vectorized HD Map elements (e.g., road topology) are fused onto the BEV feature map. The resulting hybrid representation merges background BEV perceptual context with foreground HD Map geometric structures, furnishing the VLM with high-precision structural priors.

As a comprehensive description of the surrounding driving scene, the BEV representation preserves key information about the driving environment, while filtering out data that are irrelevant to driving tasks. This characteristic significantly reduces information dimensionality, thereby improving subsequent computational efficiency. Considering that the vision-to-BEV transformation is an ill-posed problem afflicted by depth ambiguity, and pure-vision BEV feature often exhibits ray-like artifacts, we integrate LiDAR data and align them with HD Map to equip the VLM with robust spatial awareness. Meanwhile, the BEV representation serves as a unified fusion space for multimodal data, which not only addresses the limitation that VLM cannot directly leverage information from sources like LiDAR, but also enables the VLM to achieve direct fusion of diverse modalities, such as multi-sensor data and HD Map. Therefore, our work utilizes the visualized BEV-HD Map as the core input for VLM.

\subsection{Trajectory Planning}

The trajectory planning module translates the integrated BEV-HD Map into a series of precise future coordinates by leveraging the VLM’s built-in autoregressive mechanism. This process treats planning as a generalized language modeling task, where spatial waypoints are represented as special tokens.

\noindent \textbf{Geometric Alignment.} The BEV-HD Map is fed into the model as a multimodal prompt, which simply identifies the input as a unified BEV-HD Map and instructs the model to generate the upcoming waypoint. By using a top-down visual representation that naturally aligns with the ego-vehicle’s heading, the VLM can directly map the geometric feature and lane topology of the input image to spatial coordinates.

\noindent \textbf{Autoregressive Waypoint Generation.} Following the design of recent VLM for autonomous driving~\citep{yang2025less}, we formulate trajectory prediction as the sequential generation of future waypoint tokens. First, the future trajectory is tokenized into 6 discrete waypoints, each representing a \textit{0.5s} interval, resulting in a total prediction horizon of \textit{3s}. These waypoints are generated autoregressively, where each token is conditioned on both the BEV-HD Map and all previously generated tokens. Finally, the generated token sequence is decoded into physical-world coordinates in $(x,y)$ format.

\noindent \textbf{Supervision.} To ensure the accuracy of predictions, our model is supervised by the corresponding ground truth trajectory of the next \textit{3s} and optimized to minimize the prediction error using a multi-step $\ell_2$ loss:

\begin{equation}
\mathcal{L} = \sum_{t=1}^{T} \left\| \hat{\mathbf{w}}_t - \mathbf{w}_t \right\|_2^2
\end{equation}
where $T$ denotes the total number of predicted waypoints, $\hat{\mathbf{w}}_t$ represents the predicted coordinates at time step $t$, and $\mathbf{w}_t$ represents the ground truth coordinates.
\begin{figure}[!b]
    \centering
    \subfigure[BEV Map]{
        \includegraphics[width=0.3\linewidth]{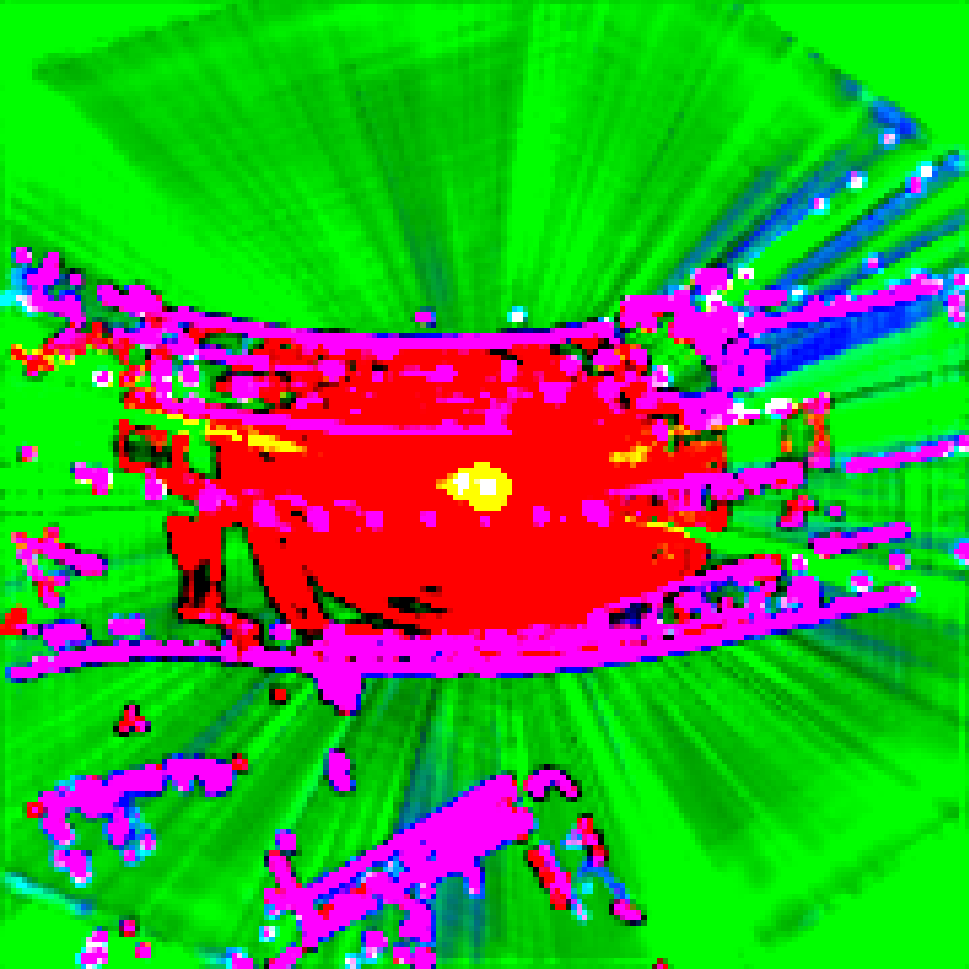}
        \label{fig:sub1}
    }
   \hspace{1pt}
    \subfigure[HD Map]{%
        \includegraphics[width=0.3\linewidth]{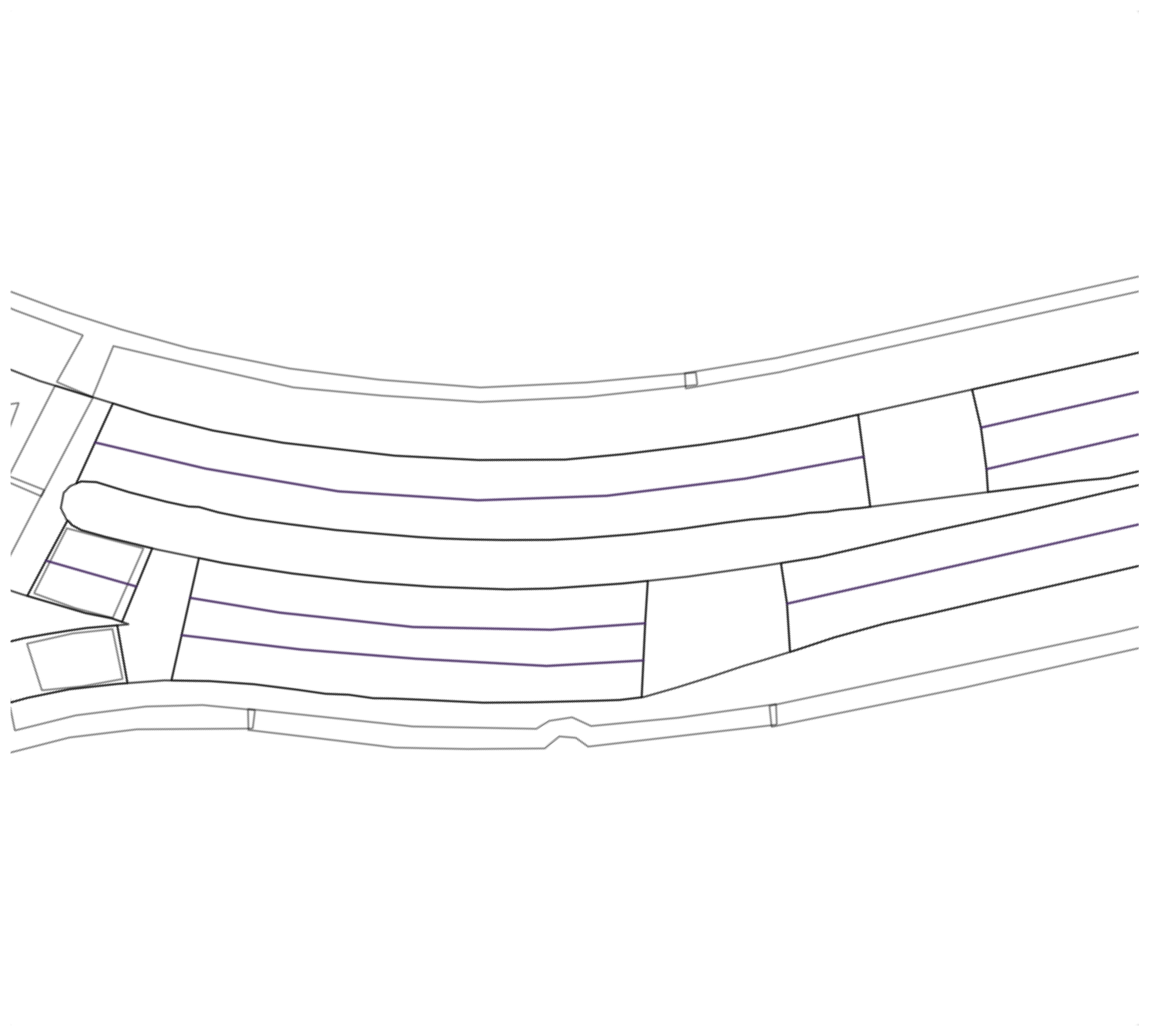}
        \label{fig:sub2}
    }
    \hspace{1pt}
    \subfigure[BEV-HD Map]{%
        \includegraphics[width=0.3\linewidth]{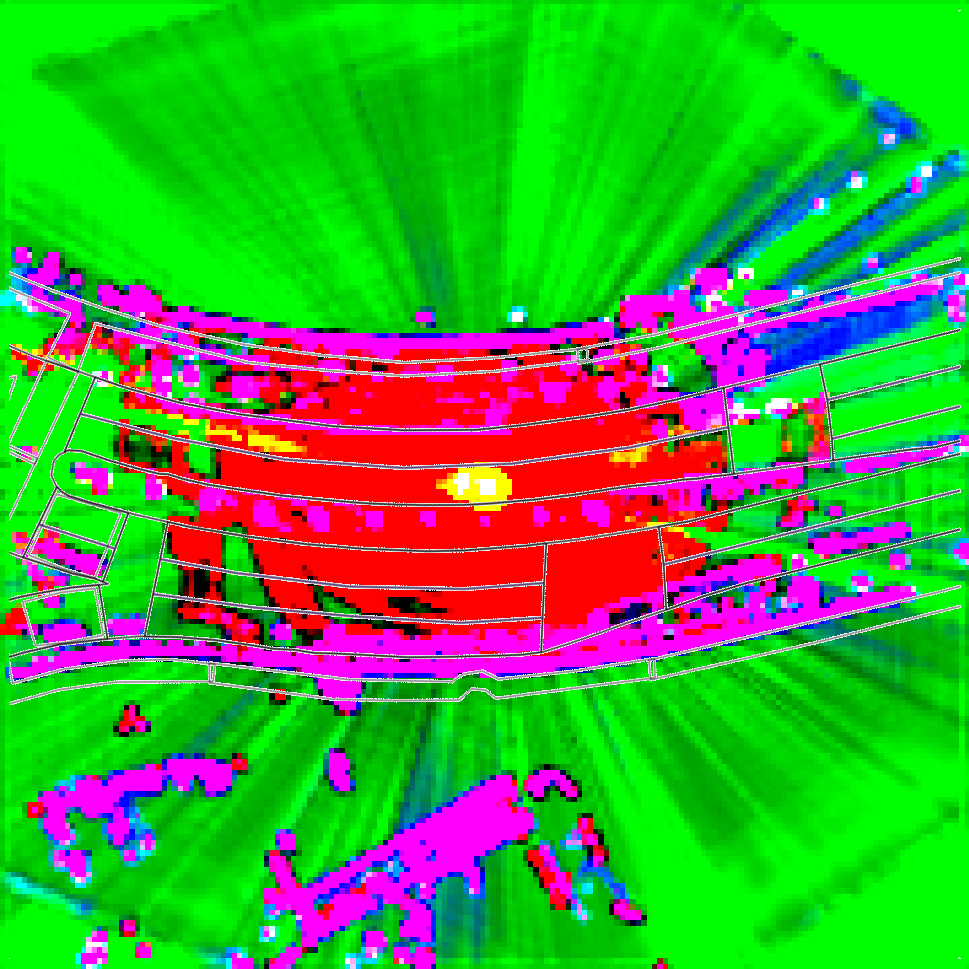}
        \label{fig:sub3}
    }
    \caption{This figure show (a) the visualized BEV Feature Map, (b) the spatially aligned HD  Map, and (c) the BEV-HD Map with precise integration rendering.}
    \label{fig:three_figures}
\end{figure}
\section{Experiments}
\subsection{Experimental Settings}

\textbf{Datasets.} In this work, we train our model on the \texttt{train} split of the nuScenes~\citep{nuscenes2019} dataset and evaluate it on the \texttt{val} split. The whole dataset, comprising over $1,000$ driving scenes each lasting approximately $20$ seconds, provides a diverse range of sensor data, including surround-view imagery, LiDAR point clouds, and more. It also offers rich annotations, such as 3D bounding boxes, ego-vehicle trajectories, and road topology maps, all of which support comprehensive model training and evaluation for autonomous driving tasks.

\noindent \textbf{Foundation Model.} We adopt \texttt{Qwen2.5-VL-3B/7B} from the Qwen-family~\citep{Qwen2.5-VL} as the foundation model.

\noindent \textbf{Prepocessing.} Data preprocessing, including prompt design and ground truth trajectory extraction, proceeds as follows: LiDAR data, images, and fused BEV feature, which cover a spatial range of $\pm50$m in both $x$, $y$ directions, are extracted from nuScenes through the pretrained BEVFusion, and then converted to the visualized RGB format through dimensionality reduction. Trajectories are extracted from nuScenes by computing ego-vehicle position differences between the current and subsequent frames.

\noindent \textbf{Metrics.} We employ mainstream $L2$ displacement error and collision rate as evaluation metrics. $L2_{\text{avg}}$ refers to displacement error under the ST-P3 standard, while $L2_{\max}$ refers to the displacement error under the UniAD standard.

\noindent \textbf{Implementation Details.} Training was conducted for $10$ epochs on $8$ NVIDIA A100 GPUs with a batch size of $1$ and a learning rate of $1\times10^{-4}$ using the AdamW optimizer. Inference was done on a single NVIDIA A100 GPU. We adopted the official BEVFusion model with standard nuScenes settings for BEV feature generation to show that even the vanilla perception module can unlock the model’s potential to leverage multimodal information within a unified BEV space.
\subsection{Main Results}
\begin{table*}[htbp]
   \centering
   \caption{\textbf{Main results.} Note that our inputs include surround-view vision, LiDAR, and HD Map. Pre-trained weights are obtained from ModelScope Platform. The \textit{Avg.} column denotes the average of the first \textit{3s}. In the following table, for Max-V1, we choose its \texttt{Qwen2.5-VL-7B} version for fair comparison. In contrast to the other models that rely solely on pure visual input, our approach incorporates fused LiDAR-camera BEV feature along with HD Map.}
   \label{table1}
    \begin{tabularx}{\textwidth}{l|l|*{4}{>{\centering\arraybackslash}X}|*{4}{>{\centering\arraybackslash}X}|*{4}{>{\centering\arraybackslash}X}}
    \toprule
    \multicolumn{2}{c|}{Model} & \multicolumn{4}{c|}{$L2_{\text{avg}}$ (m) $\downarrow$} & \multicolumn{4}{c|}{$L2_\text{{max}}$ (m) $\downarrow$} & \multicolumn{4}{c}{Collision rate (\%) $\downarrow$} \\
    \cline{3-6} \cline{7-10}\cline{11-14}
    \multicolumn{2}{c|}{} & 1s   & 2s   & 3s   & \textit{Avg.}   & 1s   & 2s   & 3s   & \textit{Avg.}  & 1s   & 2s   & 3s   & \textit{Avg.}\\
    \midrule
    \multicolumn{2}{c|}{ST-P3~\citep{hu2022stp3}}    &1.33&2.11&2.90&2.11& - & - &  - & - & 0.44 & 1.08 & 3.01 & 1.51 \\ 
    \multicolumn{2}{c|}{UniAD~\citep{hu2023_uniad}}  &0.42&0.64&0.91&0.66& 0.48 & 0.96 & 1.65 & 1.03  & 0.05 & 0.17 & 0.71 & 0.31 \\ 
    \multicolumn{2}{c|}{VAD~\citep{jiang2023vad}}      & 0.41 & 0.70 & 1.05 & 0.72& - & - &  - & -  & 0.07 & 0.17 & 0.41 & 0.22 \\ 
    \multicolumn{2}{c|}{Senna~\citep{jiang2024senna}}  & 0.26 & 0.42 & 0.61 & 0.43& - & - &  - & - & 0.04 & 0.08 & 0.13 & 0.08 \\ 
    \multicolumn{2}{c|}{AutoVLA~\citep{zhou2025autovla}}  & 0.21 & 0.38 & 0.60 & 0.40 & 0.28 & 0.66& 1.16& 0.70& 0.13 & 0.18 & 0.28 &0.20 \\ 
    \multicolumn{2}{c|}{OpenDriveVLA~\citep{zhou2025opendrivevla}}  & 0.14 & 0.30 & 0.55 & 0.33 & 0.19 & 0.58& 1.24& 0.67& 0.02 & 0.18 & 0.70 &0.30 \\
    \midrule
    \multicolumn{2}{c|}{Max-V1~\citep{yang2025less}}  &  0.24 &  0.28 &  0.46 & 0.33 & 0.23 & 0.39& 0.98& 0.53& - & - & - & - \\ 
    \multicolumn{2}{c|}{EMMA~\citep{Hwang2024EMMAEM}}    & 0.14 & 0.29 & 0.54 & 0.32  & - & - &  - & - & - & - & - & - \\ 
    
    \midrule
    \multirow{2}{*}{\textbf{BEV-VLM}} & \texttt{Qwen2.5-VL-3B}&0.14&0.44&0.50&0.36& 0.24 &  0.65&  0.99&  0.63 &  0.02&0.04 & 0.10 & 0.05 \\
                            & \texttt{Qwen2.5-VL-7B}   &\textbf{0.13} & \textbf{0.16} &  \textbf{0.16} & \textbf{0.15} &
                            \textbf{0.13} & \textbf{0.16} &  \textbf{0.18} & \textbf{0.16} & \textbf{0.00} & \textbf{0.00} & \textbf{0.00} & \textbf{0.00} \\
    \bottomrule
    \end{tabularx}
\end{table*}

\textbf{Table \ref{table1}} presents the performance of BEV-VLM in trajectory planning, compared with baseline methods. It can be observed that among all competitors, BEV-VLM based on \texttt{Qwen2.5-VL-7B} achieves the best performance. The first-person perspective visualization results are presented in Figure~\ref{fig:three_trajectories_crosscol}. We attribute this performance advantage to the core VLM’s capability to fully parse the information contained in the BEV-HD Map. Note that models with larger scale seem to have a higher upper bound for understanding BEV feature map, showing greater potential in autonomous driving.

Compared to the listed \textit{state-of-the-art} vision-only methods~\cite{Hwang2024EMMAEM}, our proposed BEV-VLM achieves $53.1\%$ decrease in the displacement error. Under the ST-P3~\citep{hu2022stp3} evaluation criterion, we achieve \textit{complete collision avoidance} for all \textit{3s} future time horizons. These results show that the VLM can well capture the scene information contained in the BEV-HD Map, thus verifying the effectiveness and reliability of our method.

\begin{table}[!h]
    \centering
    \caption[TABLE 2]{\textbf{Ablations of different visual input.} This table presents a comparison of different visual input modalities including visualized BEV feature and BEV-HD Map using \texttt{Qwen2.5-VL-7B}.}
    \label{table2}
    \begin{tabular}{c c | c c c c}
      \toprule
      \multicolumn{2}{c|}{Inputs} &
      \multicolumn{4}{c}{$ L2_\text{{avg}}(m) \downarrow$} 
       \\
      \cmidrule(lr){1-2} \cmidrule(lr){3-6}
      BEV feature map & HD Map  & 1s & 2s & 3s & \textit{Avg.}  \\
      \midrule
      \checkmark&        &\textbf{0.10}  & 0.27 & 0.45& 0.27\\
      \checkmark& \checkmark      & 0.13 & \textbf{0.16} &  \textbf{0.16} & \textbf{0.15}     \\
     
      \bottomrule
    \end{tabular}
\end{table}

\begin{figure}[t]
    \centering
    
    \subfigure[BEV-VLM ]{%
        \includegraphics[width=0.47\linewidth]{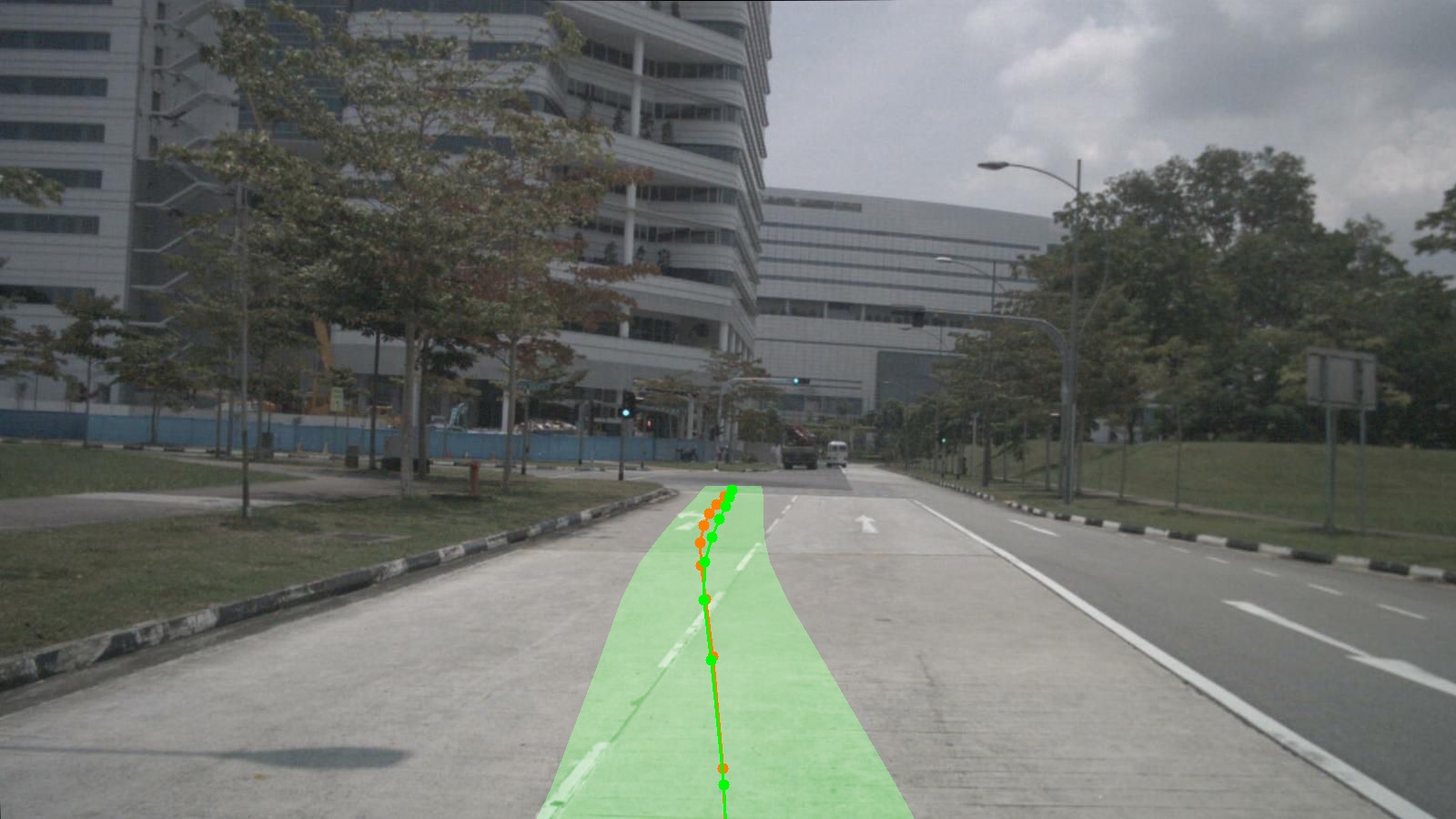}
        \label{fig:sub3}
    }
    \hspace{1pt}
    \subfigure[Max-V1]{%
        \includegraphics[width=0.47\linewidth]{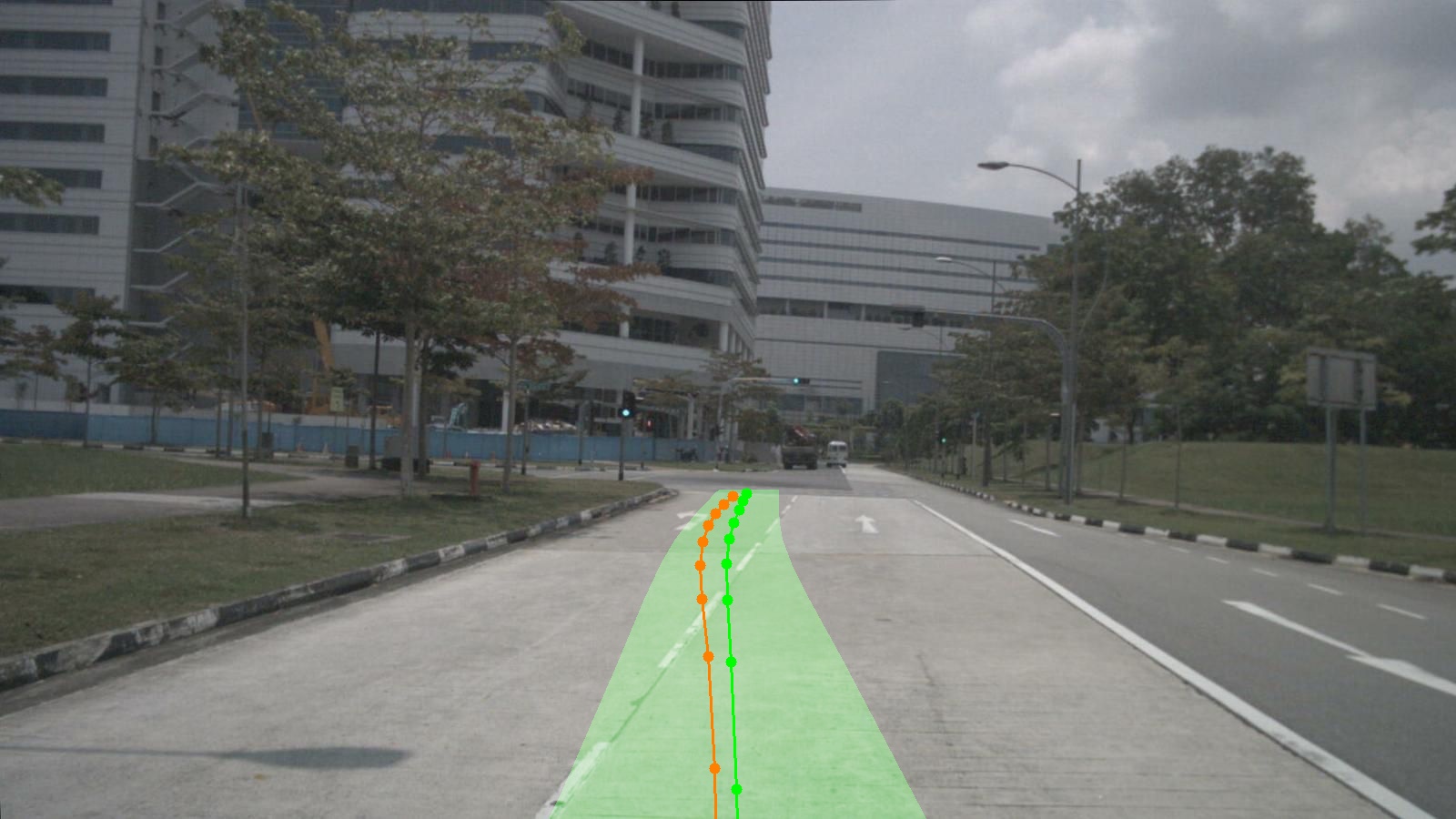}
        \label{fig:sub4}
    }
    \caption{Comparison of trajectory planning results from different input representations: trajectories generated by (a) BEV‑VLM (BEV‑HD Map), and (b) Max‑V1 (first-perspective view).}
    \label{fig4}
\end{figure}
\begin{figure}[!b]
    \centering
    \
        \includegraphics[width=1\linewidth]{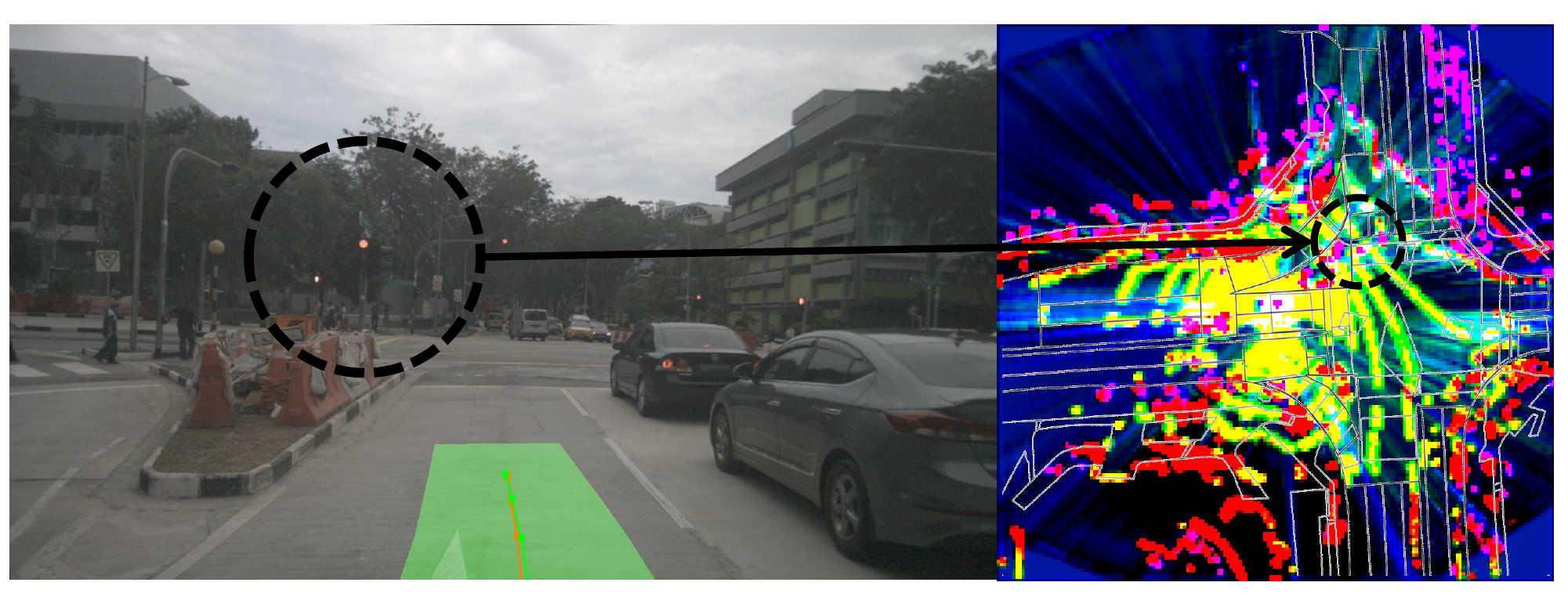}

    \caption{Front view with predicted (green) and ground truth (orange) trajectories (left), and BEV-HD Map (right). The traffic light (circled in black) and its corresponding feature are preserved in the BEV-HD Map.}
    \label{fig5}
\end{figure}
\subsection{Discussion}
We further explore the quantitative results of HD Map integration. \textbf{Table~\ref{table2}} presents the results of the ablation experiments. It can be observed that even with solely pure BEV feature without HD Map integration, the model already outperforms current \textit{state-of-the-art} methods. This further underscores the key inherent advantage of the BEV representation as a natural carrier for sensor fusion. Moreover, the performance of trajectory planning is substantially enhanced when precise topological structures from the aligned HD Map are integrated.

BEV-based representation offers additional information over perspective view-only input. While perspective views are plagued by scale ambiguity, BEV delivers a top-down metric-consistent scene layout. In this representation, each pixel is assigned a fixed distance due to the predefined coverage range; lanes are manifested as parallel structures, and ego-motion simplifies to rigid rotation. This transformation effectively converts complex 3D spatial inference into explicit 2D topological reasoning. As shown in Figure~\ref{fig4}, the perspective view struggles with accurate distance estimation when the vehicle heads leftward, while our BEV-based model preserves metric awareness, predicting physically plausible and reliable trajectories consistent with driving rules.

This geometric transformation introduces a more holistic planning paradigm. By conditioning the entire trajectory on a unified spatial context, the BEV representation effectively decouples the waypoints in time. Consequently, each waypoint is generated not merely according to its predecessor, but as part of a coherent, globally optimal path. This inherent decoupling is crucial for mitigating the compounding error problem typical of sequential, ego-view forecasting models. The integrated BEV-HD Map further supplies a global spatial context and structural priors, allowing the model to correct deviations through geometric constraints. As a result, our best-performing model shows that error accumulation over time is alleviated to some degree, which in turn supports more stable long-horizon planning.

To validate the effectiveness of our visual encoding, we measure the explained variance ratio after PCA-based compression. The results confirm that the compressed BEV feature retain $60–70\%$ of the total variance, which is demonstrated to preserve both semantically and structurally critical information. As shown in Figure~\ref{fig5}, while a red light is directly visible in the front-view image, its corresponding feature signature remains clearly discernible in the PCA-compressed BEV Map. The model successfully learns to extract this signal, yielding a trajectory that plans a full stop at the intersection. This shows that even compressed BEV representation maintains sufficient information to support the VLM in making safe and context-aware driving decisions.

\section{Conclusion}
\label{sec:conclusion}

This paper has introduced BEV-VLM, a framework that rethinks visual input for VLM in autonomous driving by shifting from raw images to unified BEV representation. By converting 3D spatial inference into 2D topological understanding, we have shown that VLM can effectively reason over geometrically consistent BEV-HD Map, which integrate multi-modal sensor data with high-definition map priors.
Our approach achieves competitive trajectory planning performance while presenting a broader paradigm. That is, VLM can robustly parse structured, task-aware visual abstraction. We harness the intrinsic merit of the BEV representation of naturally eliminating distance ambiguity to improve the spatial reasoning and reliability of the model, paving the way for safety-focused autonomous driving systems.

Future works are directed towards enriching the training data with more diverse driving scenarios and developing a unified VLM tailored to BEV feature to enhance model adaptability and simplify deployment.

\vfill\pagebreak



\bibliographystyle{IEEEbib}
\bibliography{strings,refs}

\end{document}